# Identification of the Resting Position Based on EGG, ECG, Respiration Rate and *SpO*$_2$ Using Stacked Ensemble Learning

Md. Mohsin Sarker Raihan[1], Muhammad Muinul Islam[1], Fariha Fairoz[2], and Abdullah Bin Shams[3]

[1] Department of Biomedical Engineering, Khulna University of Engineering & Technology, Khulna 9203, Bangladesh,
[2] Department of Computer Engineering, Islamic University of Technology, Gazipur, Bangladesh,
[3] Department of Electrical & Computer Engineering, University of Toronto, Toronto, Ontario M5S 3G4, Canada,
Emails: msr.raihan@gmail.com, mmi@bme.kuet.ac.bd, farihafairoz@iut-dhaka.edu, abdullahbinshams@gmail.com

**Abstract.** Rest is essential for a high-level physiological and psychological performance. It is also necessary for the muscles to repair, rebuild, and strengthen. There is a significant correlation between the quality of rest and the resting posture. Therefore, identification of the resting position is of paramount importance to maintain a healthy life. Resting postures can be classified into four basic categories: Lying on the back (supine), facing of the left / right sides and free-fall position. The later position is already considered to be an unhealthy posture by researchers equivocally and hence can be eliminated. In this paper, we analyzed the other three states of resting position based on the data collected from the physiological parameters: Electrogastrogram (EGG), Electrocardiogram (ECG), Respiration Rate, Heart Rate, and Oxygen Saturation ($SpO_2$). Based on these parameters, the resting position is classified using a hybrid stacked ensemble machine learning model designed using the Decision tree, Random Forest, and Xgboost algorithms. Our study demonstrates a 100% accurate prediction of the resting position using the hybrid model. The proposed method of identifying the resting position based on physiological parameters has the potential to be integrated into wearable devices. This is a low cost, highly accurate and autonomous technique to monitor the body posture while maintaining the user's privacy by eliminating the use of RGB camera conventionally used to conduct the polysomnography (sleep Monitoring) or resting position studies.

**Keywords:** Polysomnography, Resting Position, ECG, EGG, Stacked Ensemble Learning, Decision Tree, Random Forest, XGBoost, Machine Learning

## 1   Introduction

For physical and mental fatigue removal, human beings take rest. Despite having some similarities with sleep, rest usually indicates a shorter period and the person remains awake during rest. Lying down is one of the prominent ways of taking a rest. There are 4 basic types of resting positions. They are: Lying on the back(supine), facing either of the left or right side and free-fall position [1]. There are many insights regarding healthy sleeping or resting postures. The absence of proper rest may eventually lead to a physical and mental breakdown. Also, incorrect body posture may potentially cause muscular strain. The interrelation between the quality of rest and resting posture is quite unavoidable. Researchers are trying to determine the best-suited position by observing the volunteers using a variety of processes. Resting posture observation is considered one of the key steps in determining the cause of diverse diseases. While observing and monitoring positions, there is always a trade-off between some constraints such as privacy, use of light, cost, and the accuracy of results.

In this paper, we considered the 3 states of resting position utilizing data collected through physiological examinations. The test parameters are Electrogastrogram (EGG), Electrocardiogram (ECG), Respiration Rate, and oxygen saturation ($SpO_2$). This study used easy and simple non-invasive processes to determine resting posture. All the data are classified by using a hybrid 2-layer stacked ensemble machine learning model. Each of the layers comprises three machine learning algorithms namely Decision tree, Random Forest, and XGboost. Our findings show a significant improvement in the prediction of posture after the use of the hybrid 2-layer stacked ensemble model which is quite promising. The highest possible accuracy gain implies the credibility of the analysis and this can be used in Polysomnography afterward. Easy implementation of the insight found from this study in wearable devices is possible and reliable as the study gained high accuracy. The cost-effectiveness is also an upvote for this study.

## 2   Related Works

There is a strong relationship between sleeping or resting position with the quality of rest. Uncomfortable posture significantly diminishes the sole purpose of taking rest. There are several studies on the determination of resting and sleeping position. A large spectrum of techniques is used in this sector to get optimal performance.

Many studies have used RGB cameras which have certain limitations like the invasion of privacy, environmental noise, etc. The use of 3d model [2], Kinect [3], and other sensors are also prominent in this field. The use of a single IR camera and image classification using CNN is also found [4] to analyze posture. Some studies show the use of pressure-sensitive mats [5].

In the determination of sleeping posture, the use of pressure sensors is costly. Result calculation becomes difficult if the patient is away from the central axis. The images developed by using such sensors may lead to complex ambiguity as the number of pixels is higher. Tang et al. used sensors in a different arrangement [5]. A special mat made of force sensing registers arranging in a 2D array is used to gather statistical information about posture. Later the correct posture is determined using an artificial intelligence library "TensorFlow". The generated heat map from the sensors is used as the input and 200 images were used for each of the six positions. The accuracy of posture recognition was highest 100% in the case of an empty bed and lowest 80% for the right lateral.

Rasouli D et al. employed only one depth sensor on a tripod for the determination of posture. The study also focused on the hand and leg positions in detail. The depth signals were processed using fast Fourier Transformation on scan planes [3] 14 volunteers were involved in the experiment. The extracted features are ranked by the T-test method. The machine learning algorithm that is used is the Support Vector Machine. Postures were divided into two main groups: side and supine. This experiment also considered the scenario with and without a blanket. The average accuracy is 97.96% and the number of data to train the model was 1171.

Mohammadi et al. studied 12 positions using IR images and neural networks [4]. InfraRed images were captured using Microsoft Kinect depth camera but for simplicity depth data was removed. Later the 2D images were used for further analysis. CNN is used to classify the positions from 2D images. 10368 frames were used to train the model. The average success rate of the method was 76% with a blanket and 91% without a blanket. This study aimed at recognizing 12 positions.

## 3  Methodology

The determination of the resting position from different physiological signals instead of direct visualization or video monitoring approach may be separated into different segments as shown in Fig 1. The details of the methodology is described as follows:

### 3.1  Physiological data Collection

This study used specific physiological data and each of them was collected using reliable and sophisticated technologies. All data were recorded at once in the morning hours after a light refreshment. Allocated time for every one of the positions was 15 minutes with 2 minutes interval in between for recheck the electrode placement and subject stabilization. The data collection procedure is thoroughly explained in the study Raihan et al. with details [1]. Data was taken in the three position and they are right side, left side and supine side. Fig 2 shows the 3 resting positions we are considering for our study. Facing right, the supine position and facing left. The physiological signals collected were: Electrocardiogram (ECG), Electrogastrogram (EGG), Oxygen saturation ($SpO_2$), and Respiratory Rate (RR) at different resting positions. A signal acquisition system, MP150, Biopac Inc., USA, and AcqKnowledge applications mounted on a computer were used to capture all of the physiological signals. Fig.3 Schematic view of different physiological signals collection with different sensors and electrodes placement.

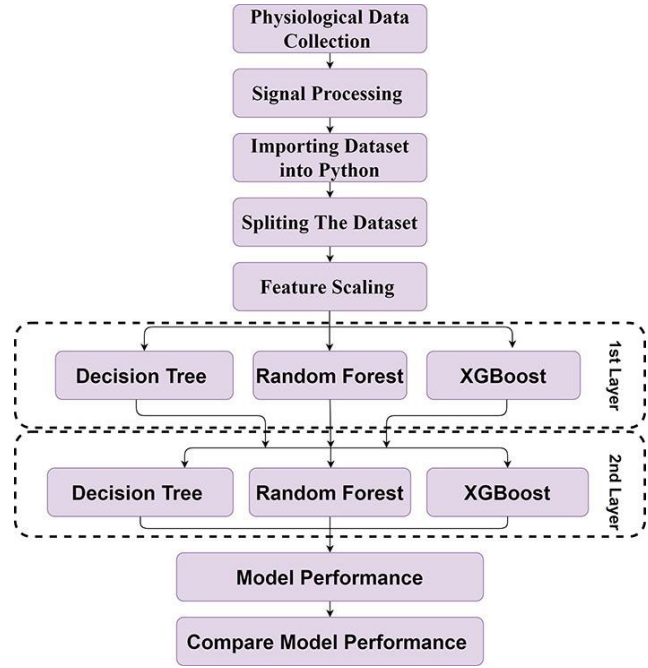

**Fig. 1.** Work-flow for the determination of the resting position from different physiological signals

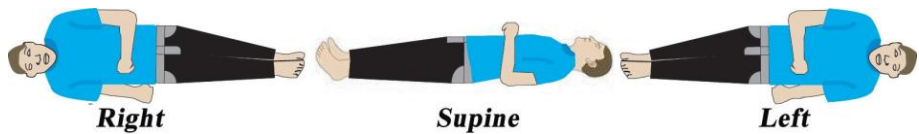

**Fig. 2.** Three types of resting positions

### 3.2   Signal Processing

EGG signal processing, Heart rate calculation procedure from ECG signal, Respiration rate and $SpO_2$ calculation methodology are followed which mention in Raihan et al [1].

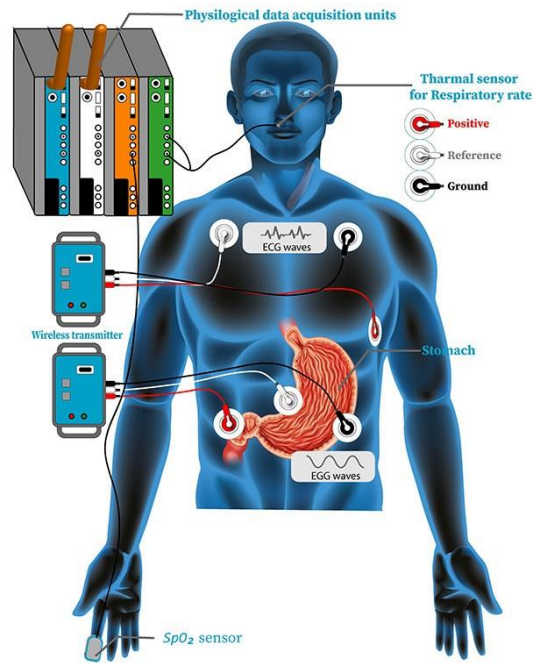

**Fig. 3.** Schematic view of different physiological signals collection with different sensors and electrode placement

### 3.3   Importing Dataset into Python

After preparation of the dataset, we imported it which consists of EGG signal, Heart rate, Respiration rate, $Sp0_2$, and resting positions.

### 3.4   Splitting the Dataset

The dataset had to be divided into two separate parts for training and testing purposes. 80% of data was used to train the model and the rest were used for testing [6].

### 3.5   Feature Scaling

The raw data we've gathered so far has a wide range of values, which could cause problems with calculations if they aren't normalized. If one of the features has a wide range of values while the others are measured in a smaller range, this feature will manipulate the calculation. As a result, the range of all features should be normalized so that each contributes roughly the same proportion [7].

### 3.6   Implemented Machine Learning Algorithms

We have implemented a heterogeneous stacked ensemble learning method with base estimators (Decision Tree, Random Forest and XgBoost as a 1st layer) and Decision Tree, Random Forest and XgBoost as our meta learner. Stacked Ensemble works by aggregating the results of multiple algorithms by using another algorithms [8]. Initially data is passed to multiple algorithms of our choosing, these are called base estimators. They classify directly using the dataset. The results of these base estimators are then passed to and subsequently aggregated by another algorithms which we call the meta leaner.We have designed a hybrid stacked ensemble learning algorithms with 2-layer. Three algorithms are used in both layers and the algorithms are described in short below.

**Decision tree:** A decision tree is a powerful yet simple machine learning algorithm. The basic induction method is recursively used to reach the prediction. As the name suggests, it creates a tree-like structure where branches reflect the outcome of the test and nodes denote any decision to be taken. The initial training set is tested and sub-divided into small sets based on definite parameters. This process is done repeatedly and called "recursive partitioning" [9].

**Random Forest:**  This algorithm stands out because of its less training time, higher accuracy for large datasets, and estimation of missing data. Random Forest uses a set of decision trees for the training process. The prediction is considered to be right based on the majority of the trees. Random Forest is included in the division of supervised learning [10].

**XGBoost:** XGBoost stands for eXtreme Gradient Boosting. Through several improvements on Gradient Boosting Machines, the XGboost has evolved. Such optimization in both algorithmic and system utilization made this perfect combination of higher accuracy in a shorter time and lesser computation. XGBoost approaches the system of sequential tree building using parallelized implementation.XGBoost uses the 'max-depth' parameter as specified rather than criterion first, and starts pruning trees backward. This algorithm ensures the efficient use of hardware resources by allocating internal buffers in each thread to store gradient statistics [10].

### 3.7   Analysis of Model Performance

The output gained from any model can easily be divided into four categories from the confusion matrix and they are True positive False-positive True negative False-negative.

   Usually, this total scenario is captured by a confusion matrix. Multiple parameters can be determined from further processing from the matrix and those are used to derive insights about the strength and effectiveness of a certain model.

The analysis of the machine learning algorithms is conducted here utilizing four derived standards from the confusion matrix. These are Accuracy, Recall, Precision, and F1 score [6]. The performance of the study is monitored using some additional parameters along with overall accuracies, such as F1 score, precision, and recall.

## 4    Results and Discussion

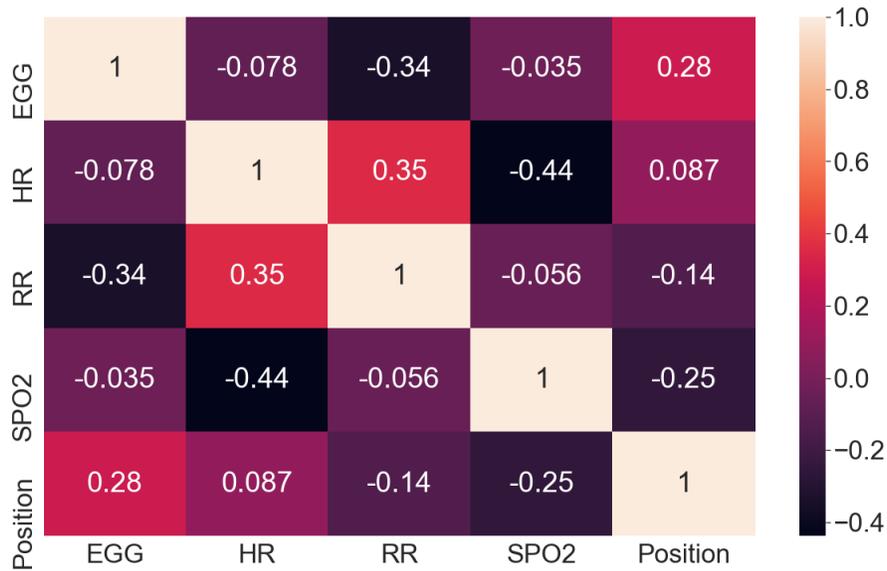

**Fig. 4.** Correlation Matrix

The correlation matrix helps to visualize the interrelationship between parameters of the dataset. It expresses the strength of variables within the range -1 to 1. Where the value close to +1 denotes a positive correlation. The closer it is to -1 it indicates the negative correlation. Neutral values or values closer to zero indicates an insignificant correlation between variables [6].

According to the matrix in Fig.4, the correlation between position and EGG is 0.28, which indicates that they have a significant positive correlation. The inverse relationship is seen for $SpO_2$ as the value is -0.25, indicating a negative correlation. Respiration rate is also inversely related to the position but not as strong as $SpO_2$. The correlation value is -0.14. Heart rate has a weak correlation with position with the value 0.087. As the value is too close to the neutral position, it implies an insignificant correlation between heart rate and position.

Fig 6 shows the results and the performance of this study. In the first layer, the accuracy of the Decision tree is 80.56% with a precision of 84.44%. The recall and F1 score were respectively 84.1% and 80.19%. While using the Random Forest, the output contained 83.33% of accuracy, precision was 81.54%, Recall 82.44% and F1 score 81.56%. For XGBoost, the accuracy was 88.89%, with a precision of 88.99%. The value of Recall and F1 score was 87.22% and 87.5% respectively.

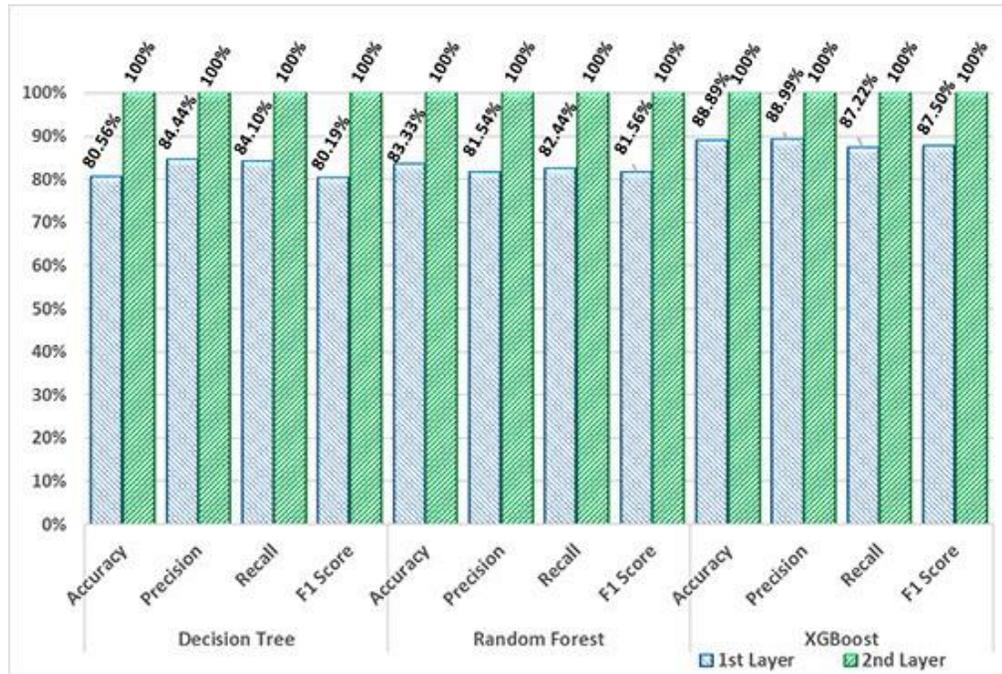

**Fig. 5.** Model Performance Result

By the end of the first layer of analysis, all the algorithms predicted the output of the training set with a percentage of accuracy as written above. Stacked ensemble learning led to higher accuracy gain. The results showed a drastic change in the second layer. In the case of stacked ensemble learning, the combination of multiple algorithms eliminates the drawbacks of each other resulting in a better gain.

As the primary predictions were collected from algorithms in the first layer, the training set along with the predictions of the previous layer was used to train the second layer. The accuracy, precision, recall, F1 score for each of the three algorithms reached 100% after the second layer of training.

**Comparison with the previous study:** If we inspect the comparison table, we will find that none of the methods mentioned in the other studies performed as well as the method mentioned in this study. In our findings we can classify with an accuracy of 100% with perfect scores in Precision, F1 and Recall. K.Tang et al [5] using SVM got 97.96% accuracy which is the nearest. Convolutional neural network with Transfer Learning got an accuracy of 90%. Other methods mentioned in the table got accuracies around 90%. It's quite evident from the table that our method shows a superior accuracy in comparison with the other methods.

**Table 1.** Comparisons with Other Existing Systems

| Ref. | Algorithm | Accuracy | Precision | F1 score | Recall |
|---|---|---|---|---|---|
| [2] | CNN with Transfer Learning | 90% | N/A | N/A | N/A |
| [5] | SVM | 97.96% | N/A | N/A | N/A |
| [10] | CNN | With Blanket:76% Without Blanket:91% | N/A | N/A | N/A |
| [11] | fuzzy c-means clustering algorithm | 88.05% | N/A | N/A | N/A |
| [12] | VGG 19 Tansor factorization | 86% | N/A | N/A | N/A |
| This study | Decision tree Random Forest XGBoost | 100% | 100% | 100% | '100% |

## 5   Conclusion

We looked at 3 resting positions using data from Electro-gastrogram (EGG), Electrocardiogram (ECG), Respiration Rate, and oxygen saturation ($SpO_2$). In our study we proposed a hybrid stacked ensemble model using Decision tree, Random Forest and Xgboost to identify the resting positon. Our method achieved an accuracy of 100%. Unlike other studies this method doesn't invade people's privacy. Also because of the high accuracy our method shows promise in being used in a cost-effective wearable device which would identify resting positions based on physiological parameters. As data was not collected via direct manual supervision, image, or video, the process was completely free from privacy violation. Such features make this study a dependable and sustainable source for future studies in a similar field of interest.

## 6   Acknowledgment

The authors wish to thank all participants during this study and cordially grateful to the Department of Biomedical Engineering, Khulna University of Engineering & Technology for proving all facilities for this study.